\documentclass{article}
\pdfoutput=1

\usepackage[preprint, nonatbib]{neurips_2019}

\usepackage[utf8]{inputenc} 
\usepackage[T1]{fontenc}    
\usepackage{hyperref}       
\usepackage{url}            
\usepackage{booktabs}       
\usepackage{amsfonts}       
\usepackage{nicefrac}       
\usepackage{microtype}      
\usepackage{times}
\usepackage{latexsym}
\usepackage{graphicx} 
\usepackage{url}
\usepackage{lipsum} 
\usepackage{multicol, blindtext}

\usepackage[font=small,skip=0pt]{caption}
\usepackage{amsmath}
\usepackage{stmaryrd}

\usepackage{amssymb}
\usepackage{hhline}
\usepackage{multirow}
\usepackage[usestackEOL]{stackengine}

\usepackage{tabularx}
\usepackage{supertabular} 

\setstackgap{C}{\normalbaselineskip}
\title{Residual Attention Net for Superior Cross-Domain Time Sequence Modeling}

\author{%
  Seth H. Huang \\
  AARC\\
  Huawei Technologies\\
  \texttt{seth.huang@huawei.com} \\
   \And
   Xu Lingjie \\
   AARC \\
   Huawei Technologies \\
    \texttt{xulingjie2@huawei.com} \\
   \And
   Jiang Congwei\\
   AARC\\
   Huawei Technologies \\
   \texttt{jiangcongwei1@huawei.com} \\
}

\begin{document}
\maketitle
\begin{abstract}
We present a novel architecture, residual attention net (RAN), which merges a sequence architecture, \textit{universal transformer}, and a computer vision architecture, \textit{residual net}, with a high-way architecture for cross-domain sequence modeling. The architecture aims at addressing the long dependency issue often faced by recurrent-neural-net-based structures. This paper serves as a proof-of-concept for a new architecture, with RAN aiming at providing the model a higher level understanding of sequence patterns. To our best knowledge, we are the first to propose such an architecture. Out of the standard 85 UCR data sets, we have achieved 35 state-of-the-art results with 10 results matching current state-of-the-art results without further model fine-tuning. The results indicate that such architecture is promising in complex, long-sequence modeling and may have vast, cross-domain applications.

\end{abstract}

\section{Introduction}

Sequence modeling and time series modeling have been critical in real life applications. Their vast applications range from market demand forecast, missile defense, weather forecast, logistics calculation, shipping routes and cost prediction to even the spreading speed of contagions. However, there have been relatively few papers addressing sequence modeling in the artificial intelligence field, currently dominated by deep learning methodologies. There is also a surging interest in the investment field to incorporate deep learning methodologies to compete in ever-competitive financial markets \cite{kim2019financial}. However, there are three main challenges facing this type of data. 

First, real life data may have long-dependency issues. The most popular sequence modeling methodology in artificial intelligence has been long-short-term memory (LSTM), which is a type of recurrent neural network. Second, financial data notoriously has high noise-to-signal ratio, which means the random information to actual pattern information can be quite imbalanced, and despite the modeling power of deep neural networks, the model learns primarily noises. It overfits easily and cannot generalize well \cite{kim2019financial}. 

Lastly, sequence data can often be non-stationary. This means that the patterns often change, and what the model learns may not be applicable to the new market scenarios. One way to compensate for this is to extend the look-back period to incorporate for historical data in the past so the model can recognize more pattern types \cite{zeng2019holistic}. This is also related to the first challenge as few models can address the long-dependency issues - though LSTM was first created to tackle this challenge, it has not been able to solve this issue effectively and has since been replaced by other attention-based algorithms especially in the natural-language domain \cite{riou2019reinforcement}. 

For sequence modeling tasks with deep learning, natural-language processing (NLP) has been at the forefront of innovations. For speech recognition and translation domains, many types of new architectures employing attention mechanism have been created such as attention-LSTM \cite{liu2017global}, transformer \cite{guo2019star}(bert), and Bert \cite{devlin2018bert} . 

\cite{ahmed2017weighted} has shown that attention mechanism has been much more effective than LSTM in language tasks, and this idea was recently adopted by \cite{song2018attend} to tackle sequence classification tasks. Compared to other modeling techniques, the authors claim attention-based methodologies are better at addressing long-dependency issues \cite{bin2018describing}. 

Additionally, some researchers have adopted techniques from the computer vision field and transfer the architectures to sequence modeling. For example, \cite{bai2018empirical} uses convolution neural net for sequence modeling. \cite{seo2018structured}. Recently in the statistics field, \cite{he2016deep} uses residual neural net (ResNet) for sequence modeling and outperforms many benchmarks based on statistical method. Some newly developed statistical methods such as Hive and Hive-Cote have shown to be extremely powerful in some sequence modeling \cite{fawaz2019deep}, but they are ensemble methods based on numerous previous models, and they take extremely long time to train \cite{shifaz2019ts} and are not practical for real-life applications \cite{shifaz2019ts}. The limitations call for a sophisticated model architecture which can be readily applied in a practical setting while alleviating the above three challenges. 

Typically, it is in theory inappropriate to use convolution-based methodology to tackle sequence modeling. Though it does not have long-dependency issues \cite{martin2019neural} as the data points are not passed through to the next block but are output directly in the pooling layer (average or global), convolution-based methodologies are "translation-invariant," which simply means time and sequence are not captured in CNN-based models - the architecture does not care where the pattern is and only cares that it appears in the sequence \cite{zhang2019linked}. This is in theory inappropriate for time-driven sequence modeling especially for noisy data. 

A recent series of papers \cite{karim2019insights,karim2017lstm,karim2019multivariate} focusing on LSTM-FCN, which combines LSTM and fully convolutional net (check proper name) and achieved numerous state-of-the-art (SOTA) results on public benchmarks. This paper series will feature prominently in our paper and has inspired our work here. 

LSTM-FCN architecture focuses on two branches, once based on LSTM (and a variant attention-LSTM, an LSTM with attention block), and the other based on a simple, three-block convolutions. For a recent analysis \cite{karim2019insights}, in what they call an "ablation test," they present the data represented in the CNN filters. The intriguing part for us is that, based on the visual examination, the filters, depending on the data sets, sometimes have a smoothing effect and other times making the sequence noisier while retaining the overall sequence structure. Though not the focus of this paper, a hypothesis is that, while the sequence model is able to capture the time aspect of the data, the CNN architecture through the filters can make the models more resistant to noises. 

Our work here elaborates on this concept and attempts to address the three main issues facing sequence modeling by combining a transformer architecture \cite{vaswani2017attention} and a residual, high-way architecture \cite{zhang2018residual} and examine the effectiveness with UCR's 85 public, sequence classification tasks. 

\section{Related Works}

\subsection{ResNet}
Residual Networks called ResNet use residuals to reconstruct the mapping of the network. That is to say, the input x is introduced again to the result, so that the weight of the stacked layer tends to zero.
\begin{equation}
 y=\mathcal{F}(x,\{W_i\})+x
\end{equation}

Here $x$ and $y$ are the input and output vectors of the layers considered. The function $\mathcal{F}(x,\{W_i\})$ is the
residual mapping to be learned. 

In order to match the dimension of $\mathcal{F}(x,\{W_i\})$ and $x$, add a linear projection $W_s$ to the above equation as:
\begin{equation}
 y=\mathcal{F}(x,\{W_i\})+W_sx
\end{equation}

\subsection{Universal Transformer}
The Universal Transformer,  which is based on encoder-decoder architecture, has been proposed by \cite{dehghani2018universal}. The details of this architecture is briefly shown as follows.

\textbf{Encoder:} 
At first, for an input sequence with length $m$, initialize a matrix $H^0 \in \mathbb{R}^{ m \times d}$, and each row of it is a $d$-dimensional embedding of the symbols at each position of the sequence. Then use the multi-headed dot-product self-attention mechanism to compute representations $H^t$ at step $t$, followed by a recurrent transition function. In addition, output and layer normalization are also added as the residual connections. 

Use the scaled dot-product attention which combines queries $Q$, keys $K$ and values $V$ as follows 
\begin{equation}
 ATTENTION(Q,K,V)=SOFTMAX(\frac{QK^T}{\sqrt{d}})V
\end{equation}

$d$ is the number of columns of $Q$, $K$ and $V$. Then based on this equation, use the multi-head version with $k$ heads.

\begin{equation}
MUITIHEADSELFATTENTION(H^t)=CONCAT(head_1,...,head_k)W^O
\end{equation}
where
\begin{equation}
head_1=ATTENTION(H^tW_i^Q,H^tW_i^K,H^tW_i^V)
\end{equation}
and $W^Q, W^K, W^V, W^O\in \mathbb{R}^{ d \times d/k}$.

So the revised representations $H^t \in \mathbb{R}^{ m \times d}$ are computed as follows
\begin{equation}
H^t=LAYERNORM(A^t+TRANSITION(A^t))
\end{equation}
where
\begin{equation}
A^t=LAYERNORM((H^{t-1}+P^t)+MULTIHEADAELFATTENTION(H^{t-1}+P^t))
\end{equation}

$P^t \in \mathbb{R}^{m \times d}$ are fixed, constant, two-dimensional (position, time) coordinate embeddings, for the positions $1 \leq i \leq m$ and the time-step $1 \leq t \leq T$ separately for each vector-dimension $1 \leq j \leq d$, 
\begin{equation}
P_{i,2j}^t=\sin{i/10000^{2j/d}}+\sin{t/10000^{2j/d}}
\end{equation}
\begin{equation}
P_{i,2j+1}^t=\cos{i/10000^{2j/d}}+\cos{t/10000^{2j/d}}.
\end{equation}

Finally, the output of encoder  is a matrix of d-dimensional vector representation $H^T \in \mathbb{R}^{ m \times d}$ for the $m$symbols of the input sequence after $T$ steps\cite{dehghani2018universal}.

\textbf{Decoder:}
The structure of the decoder is the same as the encoder. Specifically, after the self-attention function, the decoder also attends to the final encoder representation  $H^T$ of each position in input sequence using the Equation($2$). But the queries $Q$ obtained from projecting the decoder representation, and keys and values obtained from projecting the encoder representation\cite{dehghani2018universal}.
During training, the decoder input is the target output, and finally, per-symbol target distributes are obtained: 
\begin{equation}
P(y_{pos}|y_{[1:pos-1]},H^T)=SOFTMAX(OH^T)
\end{equation}
where
$O \in \mathbb{R}^d \times V$ is an affine transformation from the final decoder state to the output vacabulary size $V$. Softmax yields an $(m \times V)$-dimension output matrix normalized over its rows. Then select the maximal probability symbol as the next symbol.

\begin{figure}[!htb] 
\centering 
\includegraphics[width=1\textwidth]{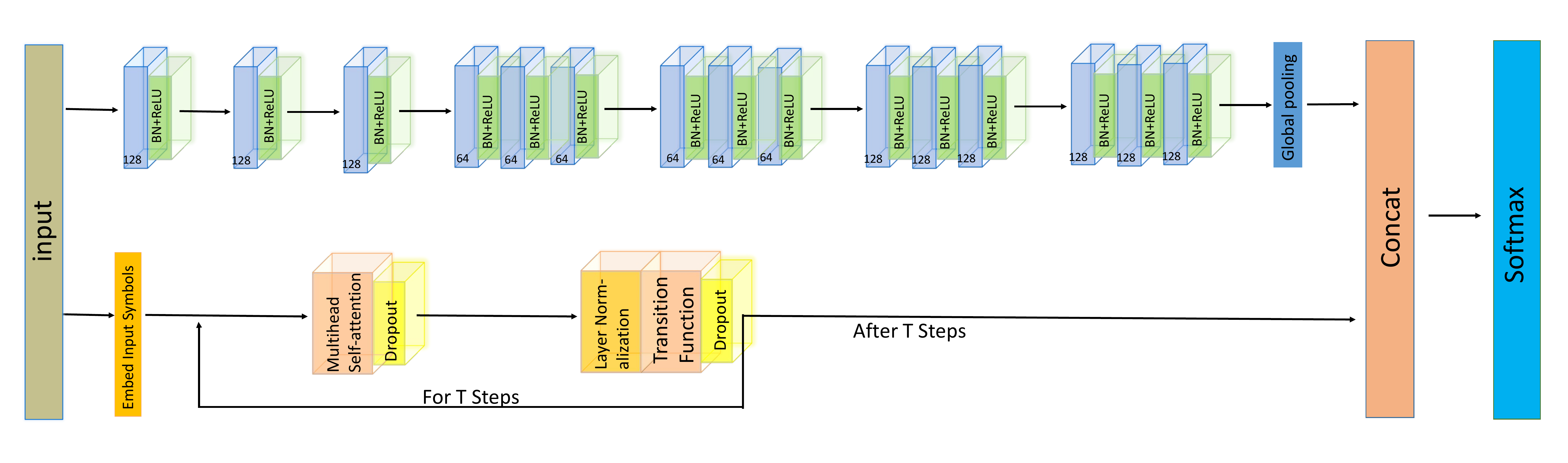} 
\caption{Resnet-Transformer architecture} 
\label{Fig.main2} 
\end{figure}

Figure\ref{Fig.main2} is just the architecture of the combination of ResNet and Universal Transformer.


\section{Experimental Results}

Here, we present the experimental results between previously existing SOTAs, the LSTM-FCN results (in some cases exceeding the SOTAs) and the Resnet-Transformer results. For LSTM-FCN, we reproduced experimental results and took the best results among several training cycles, following the approach in \cite{karim2019insights}. Notice that, the reported results in \cite{karim2019insights} are fine-tuned for each data set, i.e. the model was adjusted to achieve the best results in individual datasets. The practicality of fine-tuning to boost the validation/ out-of-sample data set performance may produce over-fitted models when not combined with ensemble methods. we tried to achieve similar effects by varying the size of fully-convolutional layers and then took the best results. 

To compare, we provide the varied architectures by varying the depth size of the transformer branch and the ResNet feature maps.  

\begin{center}

\small{
 
\captionof{table}{Performance comparison of proposed models with the rest}
 \tablefirsthead{%
  
    \hline
    \multicolumn{1}{|c|}{\multirow{5}[10]{*}{dataset\_name}} & \multicolumn{1}{c|}{\multirow{5}[10]{*}{Existing SOTA}} & \multicolumn{1}{c|}{\multirow{5}[10]{*}{Best:\newline{}lstm-fcn}} & \multicolumn{1}{c|}{\multirow{5}[10]{1.6cm}{Vanilla:\newline{}ResNet-Transformer}} & \multicolumn{3}{p{15em}<{\centering}|}{ResNet-Transformer} \\
    
    \hhline{|~~~~---|}        &       &       &       & \multicolumn{3}{p{15em}<{\centering}|}{Transformer depth} \\
   
    \hhline{|~~~~---|}          &       &       &       & 1     & 4     & 4 \\
   
    \hhline{|~~~~---|}          &       &       &       & \multicolumn{3}{p{15em}<{\centering}|}{ResNet feature maps} \\
   
    \hhline{|~~~~---|}          &       &       &       & \multicolumn{1}{p{5em}<{\centering}|}{[128, 128, 64, 64]} & \multicolumn{1}{p{5em}<{\centering}|}{[128, 128, 64, 64]} & \multicolumn{1}{p{5em}<{\centering}|}{[64, 64, 128, 128]} \\
    }
  \tablehead{%
    \hline
    \multicolumn{7}{|l|}{\small\sl continued from previous page}\\
    \hline
    \multicolumn{1}{|c|}{\multirow{5}[10]{*}{dataset\_name}} & \multicolumn{1}{c|}{\multirow{5}[10]{*}{Existing SOTA}} & \multicolumn{1}{c|}{\multirow{5}[10]{*}{Best:\newline{}lstm-fcn}} & \multicolumn{1}{c|}{\multirow{5}[10]{1.6cm}{Best:\newline{}ResNet-Transformer}} & \multicolumn{3}{p{15em}<{\centering}|}{ResNet-Transformer} \\
  
    \hhline{|~~~~---|}          &       &       &       & \multicolumn{3}{p{15em}<{\centering}|}{Transformer depth} \\
    
    \hhline{|~~~~---|}          &       &       &       & 1     & 4     & 4 \\
    \hhline{|~~~~---|}          &       &       &       & \multicolumn{3}{p{15em}<{\centering}|}{ResNet feature maps} \\
   
    \hhline{|~~~~---|}          &       &       &       & \multicolumn{1}{p{5em}<{\centering}|}{[128, 128, 64, 64]} & \multicolumn{1}{p{5em}<{\centering}|}{[128, 128, 64, 64]} & \multicolumn{1}{p{5em}<{\centering}|}{[64, 64, 128, 128]} \\
   }
  \tabletail{%
    \hline
    \multicolumn{7}{|r|}{\small\sl continued on next page}\\
    \hline}
  \tablelasttail{\hline}

   \setlength{\tabcolsep}{0.5pt}
   
    \begin{supertabular}{|l|c|c|c|c|c|c|}

    \hline
  

    Adiac & 0.857 & \textbf{0.869565} & 0.84399 & 0.849105 & 0.849105 & 0.849105 \\
    \hline

    ArrowHead & 0.88  & \textbf{0.925714} & 0.891429 & 0.891429 & 0.891429 & 0.897143 \\
    \hline

    ChlorineConcentration & \textbf{0.872} & 0.816146 & 0.849479 & 0.863281 & 0.409375 & 0.861719 \\
    \hline
    InsectWingbeatSound & 0.6525 & \textbf{0.668687} & 0.522222 & 0.642424 & 0.535859 & 0.536364 \\
    \hline
    Lightning7 & 0.863 & \textbf{0.863014} & 0.821918 & 0.849315 & 0.383562 & 0.835616 \\
    \hline
    Wine  & 0.889 & 0.833333 & 0.851852 & 0.87037 & 0.87037 & \textbf{0.907407} \\
    \hline
    WordSynonyms & \textbf{0.779} & 0.680251 & 0.661442 & 0.65047 & 0.636364 & 0.678683 \\
    \hline
    Beef  & \textbf{0.9} & \textbf{0.9} & 0.866667 & 0.866667 & 0.866667 & 0.866667 \\
    \hline
    DistalPhalanxOutlineAgeGroup & \textbf{0.835} & 0.791367 & 0.81295 & 0.776978 & 0.467626 & 0.776978 \\
    \hline
    DistalPhalanxOutlineCorrect & 0.82  & 0.797101 & \textbf{0.822464} & \textbf{0.822464} & \textbf{0.822464} & 0.793478 \\
    \hline
    DistalPhalanxTW & \textbf{0.79} & 0.748201 & 0.733813 & 0.748201 & 0.719424 & 0.741007 \\
    \hline
    ECG200 & 0.92  & 0.91  & 0.94  & \textbf{0.95} & 0.94  & 0.93 \\
    \hline
    ECGFiveDays & \textbf{1} & 0.987224 & \textbf{1} & \textbf{1} & \textbf{1} & \textbf{1} \\\hline
    BeetleFly & 0.95  & \textbf{1} & \textbf{1} & 0.95  & 0.95  & \textbf{1} \\
    \hline
    BirdChicken & 0.95  & 0.95  & \textbf{1} & 0.9   & \textbf{1} & 0.7 \\
    \hline
    ItalyPowerDemand & 0.97  & 0.963071 & 0.965015 & 0.969874 & 0.962099 & \textbf{0.971817} \\
    \hline
    SonyAIBORobotSurface1 & 0.985 & 0.985025 & \textbf{0.988353} & 0.978369 & 0.708819 & 0.985025 \\
    \hline
    SonyAIBORobotSurface2 & 0.962 & 0.972718 & 0.976915 & 0.974816 & \textbf{0.98426} & 0.976915 \\
    \hline
    MiddlePhalanxOutlineAgeGroup & \textbf{0.8144} & 0.668831 & 0.655844 & 0.662338 & 0.623377 & 0.662338 \\
    \hline
    MiddlePhalanxOutlineCorrect & 0.8076 & 0.841924 & \textbf{0.848797} & \textbf{0.848797} & \textbf{0.848797} & 0.835052 \\
    \hline
    MiddlePhalanxTW & 0.612 & 0.603896 & 0.564935 & 0.577922 & 0.551948 & \textbf{0.623377} \\
    \hline
    ProximalPhalanxOutlineAgeGroup & 0.8832 & 0.887805 & 0.887805 & \textbf{0.892683} & 0.882927 & \textbf{0.892683} \\
    \hline
    ProximalPhalanxOutlineCorrect & 0.918 & \textbf{0.931271} & \textbf{0.931271} & \textbf{0.931271} & 0.683849 & 0.924399 \\
    \hline
    ProximalPhalanxTW & 0.815 & \textbf{0.843902} & 0.819512 & 0.814634 & 0.819512 & 0.819512 \\
    \hline
    MoteStrain & \textbf{0.95} & 0.938498 & 0.940895 & 0.916933 & 0.9377 & 0.679712 \\
    \hline
    MedicalImages & 0.792 & \textbf{0.798684} & 0.780263 & 0.765789 & 0.759211 & 0.789474 \\
    \hline
    Strawberry & 0.976 & \textbf{0.986486} & \textbf{0.986486} & \textbf{0.986486} & \textbf{0.986486} & \textbf{0.986486} \\
    \hline
    ToeSegmentation1 & 0.9737 & \textbf{0.991228} & 0.969298 & 0.969298 & 0.97807 & \textbf{0.991228} \\
    \hline
    Coffee & \textbf{1} & \textbf{1} & \textbf{1} & \textbf{1} & \textbf{1} & \textbf{1} \\
    \hline
    CricketX & 0.821 & 0.792308 & \textbf{0.838462} & 0.8   & 0.810256 & 0.8 \\
    \hline
    CricketY & 0.8256 & 0.802564 & \textbf{0.838462} & 0.820513 & 0.825641 & 0.807692 \\
    \hline
    CricketZ & 0.8154 & 0.807692 & \textbf{0.820513} & 0.805128 & 0.128205 & 0.1 \\
    \hline
    UWaveGestureLibraryX & 0.8308 & \textbf{0.843663} & 0.780849 & 0.814629 & 0.810999 & 0.808766 \\
    \hline
    UWaveGestureLibraryY & 0.7585 & \textbf{0.765215} & 0.664992 & 0.71636 & 0.671413 & 0.67895 \\
    \hline
    UWaveGestureLibraryZ & 0.7725 & \textbf{0.795924} & 0.756002 & 0.761027 & 0.760469 & 0.762144 \\
    \hline
    ToeSegmentation2 & 0.9615 & 0.930769 & \textbf{0.976923} & 0.953846 & 0.953846 & \textbf{0.976923} \\
    \hline
    DiatomSizeReduction & 0.967 & 0.970588 & 0.993464 & \textbf{0.996732} & 0.379085 & \textbf{0.996732} \\
    \hline
    car   & 0.933 & \textbf{0.966667} & 0.95  & 0.883333 & 0.866667 & 0.3 \\
    \hline
    CBF   & \textbf{1} & 0.996667 & \textbf{1} & 0.997778 & \textbf{1} & \textbf{1} \\
    \hline
    CinCECGTorso & \textbf{0.9949} & 0.904348 & 0.871739 & 0.656522 & 0.89058 & 0.31087 \\
    \hline
    Computers & 0.848 & 0.852 & 0.86  & 0.844 & \textbf{0.908} & 0.84 \\
    \hline
    Earthquakes & 0.801 & \textbf{0.81295} & 0.755396 & 0.755396 & 0.76259 & 0.755396 \\
    \hline
    ECG5000 & 0.9482 & \textbf{0.948222} & 0.941556 & 0.943556 & 0.944222 & 0.940444 \\
    \hline
    ElectricDevices & \textbf{0.7993} & 0.779665 & 0.774219 & 0.771625 & 0.757489 & 0.766178 \\
    \hline
    FaceAll & 0.929 & \textbf{0.956213} & 0.881065 & 0.848521 & 0.949704 & 0.252071 \\
    \hline
    FaceFour & \textbf{1} & 0.943182 & 0.954545 & 0.965909 & 0.977273 & 0.215909 \\
    \hline
    FacesUCR & \textbf{0.958} & 0.941463 & 0.957561 & 0.947805 & 0.926829 & 0.95122 \\
    \hline
    Fish  & 0.989 & 0.971429 & \textbf{1} & 0.977143 & 0.96  & 0.994286 \\
    \hline
    FordA & 0.9727 & \textbf{0.976515} & 0.948485 & 0.946212 & 0.517424 & 0.940909 \\
    \hline
    FordB & \textbf{0.9173} & 0.792593 & 0.838272 & 0.830864 & 0.838272 & 0.823457 \\
    \hline
    GunPoint & \textbf{1} & \textbf{1} & \textbf{1} & \textbf{1} & \textbf{1} & \textbf{1} \\
    \hline
    Ham   & 0.781 & \textbf{0.809524} & 0.761905 & 0.780952 & 0.619048 & 0.514286 \\
    \hline
    HandOutlines & 0.9487 & \textbf{0.954054} & 0.937838 & 0.948649 & 0.835135 & 0.945946 \\
    \hline
    Haptics & 0.551 & 0.558442 & 0.564935 & 0.545455 & \textbf{0.600649} & 0.194805 \\
    \hline
    Herring & 0.703 & \textbf{0.75} & 0.703125 & 0.734375 & 0.65625 & 0.703125 \\
    \hline
    InlineSkate & \textbf{0.6127} & 0.489091 & 0.516364 & 0.494545 & 0.494545 & 0.165455 \\
    \hline
    LargeKitchenAppliances & 0.896 & 0.898667 & 0.928 & 0.898667 & \textbf{0.936} & 0.933333 \\
    \hline
    Lightning2 & \textbf{0.8853} & 0.819672 & 0.852459 & 0.852459 & 0.754098 & 0.868852 \\
    \hline
    MALLAT & 0.98  & \textbf{0.98081} & 0.977399 & 0.975267 & 0.934328 & 0.979104 \\
    \hline
    Meat  & \textbf{1} & 0.883333 & \textbf{1} & \textbf{1} & \textbf{1} & \textbf{1} \\
    \hline
    NonInvasiveFetalECGThorax1 & 0.961 & \textbf{0.970483} & 0.953181 & 0.953181 & 0.947583 & 0.948092 \\
    \hline
    NonInvasiveFetalECGThorax2 & 0.955 & \textbf{0.961323} & 0.955216 & 0.954198 & 0.948601 & 0.952672 \\
    \hline
    OliveOil & 0.9333 & 0.766667 & \textbf{0.966667} & 0.9   & 0.933333 & 0.9 \\
    \hline
    OSULeaf & 0.988 & 0.983471 & 0.987603 & \textbf{0.991736} & 0.987603 & \textbf{0.991736} \\
    \hline
    PhalangesOutlinesCorrect & 0.83  & 0.83683 & \textbf{0.855478} & 0.848485 & 0.854312 & 0.850816 \\
    \hline
    Phoneme & 0.3492 & 0.341772 & \textbf{0.363924} & 0.191983 & 0.357595 & 0.348101 \\
    \hline
    plane & \textbf{1} & \textbf{1} & \textbf{1} & \textbf{1} & 0.371429 & \textbf{1} \\
    \hline
    RefrigerationDevices & 0.5813 & 0.605333 & 0.605333 & 0.616 & 0.592 & \textbf{0.618667} \\
    \hline
    ScreenType & \textbf{0.707} & 0.682667 & 0.669333 & 0.645333 & 0.666667 & 0.68 \\
    \hline
    ShapeletSim & \textbf{1} & \textbf{1} & \textbf{1} & 0.911111 & 0.888889 & 0.977778 \\
    \hline
    ShapesAll & 0.9183 & 0.905 & 0.923333 & 0.876667 & 0.921667 & \textbf{0.933333} \\
    \hline
    SmallKitchenAppliances & 0.803 & 0.821333 & 0.808 & 0.810667 & \textbf{0.829333} & 0.813333 \\
    \hline
    StarlightCurves & 0.9796 & 0.977295 & \textbf{0.978873} & 0.979237 & \textbf{0.978873} & 0.975838 \\
    \hline
    SwedishLeaf & 0.9664 & \textbf{0.9792} & \textbf{0.9792} & 0.9728 & 0.9696 & 0.9664 \\
    \hline
    Symbols & 0.9668 & \textbf{0.98794} & 0.9799 & 0.970854 & 0.976884 & 0.252261 \\
    \hline
    SyntheticControl & \textbf{1} & 0.993333 & \textbf{1} & 0.996667 & \textbf{1} & \textbf{1} \\
    \hline
    Trace & \textbf{1} & \textbf{1} & \textbf{1} & \textbf{1} & \textbf{1} & \textbf{1} \\
    \hline
    TwoPatterns & \textbf{1} & 0.99675 & \textbf{1} & \textbf{1} & \textbf{1} & \textbf{1} \\
    \hline
    TwoLeadECG & \textbf{1} & \textbf{1} & \textbf{1} & \textbf{1} & \textbf{1} & \textbf{1} \\
    \hline
    UWaveGestureLibraryAll & \textbf{0.9685} & 0.961195 & 0.856784 & 0.933277 & 0.939978 & 0.879118 \\
    \hline
    wafer & \textbf{1} & 0.998378 & 0.99854 & 0.998215 & 0.99854 & 0.999027 \\
    \hline
    Worms & 0.8052 & \textbf{0.844156} & 0.831169 & 0.779221 & 0.818182 & 0.25974 \\
    \hline
    WormsTwoClass & 0.8312 & 0.844156 & \textbf{0.857143} & 0.831169 & 0.805195 & 0.831169 \\
    \hline
    yoga  & 0.9183 & \textbf{0.921667} & 0.906333 & 0.905667 & 0.884 & 0.866667 \\
    \hline
    
    ACSF1 & -    & 0.9   & \textbf{0.96} & 0.91  & 0.93  & 0.17 \\
    \hline
    AllGestureWiimoteX & -      & 0.701429 & 0.77  & 0.76  & \textbf{0.762857} & 0.754286 \\
    \hline
    AllGestureWiimoteY & -     & 0.802857 & \textbf{0.814286} & 0.798571 & 0.808571 & 0.8 \\
    \hline
    AllGestureWiimoteZ &  -     & 0.684286 & \textbf{0.782857} & 0.752857 & 0.767143 & 0.748571 \\
    \hline
    BME   &  -     & 0.993333 & \textbf{1} & \textbf{1} & \textbf{1} & \textbf{1} \\
    \hline
    Chinatown &  -     & 0.982609 & \textbf{0.985507} & \textbf{0.985507} & \textbf{0.985507} & \textbf{0.985507} \\
    \hline
    Crop  &  -     & 0.74494 & 0.743869 & 0.742738 & \textbf{0.746012} & 0.740476 \\
    \hline
    DodgerLoopDay &  -     & \textbf{0.6375} & 0.5375 & 0.55  & 0.4625 & 0.5 \\
    \hline
    DodgerLoopGame & -      & 0.898551 & 0.876812 & 0.891304 & 0.550725 & \textbf{0.905797} \\
    \hline
    DodgerLoopWeekend &  -     & \textbf{0.978261} & 0.963768 & 0.978261 & 0.949275 & 0.963768 \\
    \hline
    EOGHorizontalSignal &  -     & \textbf{0.654696} & 0.610497 & 0.59116 & 0.60221 & 0.610497 \\
    \hline
    EOGVerticalSignal &  -     & \textbf{0.505525} & 0.450276 & 0.48895 & 0.146409 & 0.480663 \\
    \hline
    EthanolLevel & -      & 0.772 & 0.824 & \textbf{0.868} & 0.82  & 0.802 \\
    \hline
    FreezerRegularTrain & -      & 0.997895 & \textbf{0.999649} & 0.999298 & 0.999298 & \textbf{0.999649} \\
    \hline
   
    FreezerSmallTrain &  -     & 0.825965 & \textbf{0.975088} & 0.958947 & 0.906667 & 0.771579 \\
    \hline
   
    Fungi & -      & 0.994624 & \textbf{1} & \textbf{1} & 0.994624 & 0.075269 \\
    \hline
   
    GestureMidAirD1 &   -    & 0.715385 & 0.715385 & \textbf{0.723077} & \textbf{0.723077} & 0.7 \\
    \hline

    GestureMidAirD2 &  -     & 0.707692 & \textbf{0.746154} & 0.692308 & 0.676923 & 0.7 \\
    \hline
   
    GestureMidAirD3 & -      & \textbf{0.430769} & 0.353846 & 0.369231 & 0.338462 & 0.338462 \\
    \hline
 
    GesturePebbleZ1 &  -     & 0.918605 & \textbf{0.936047} & 0.831395 & \textbf{0.936047} & 0.906977 \\
    \hline
   
    GesturePebbleZ2 &  -     & 0.886076 & 0.873418 & 0.841772 & \textbf{0.911392} & 0.879747 \\
    \hline

    GunPointAgeSpan & -      & 0.996835 & 0.996835 & 0.996835 & \textbf{1} & 0.848101 \\
    \hline
   
    GunPointMaleVersusFemale &  -     & \textbf{1} & \textbf{1} & \textbf{1} & 0.996835 & 0.996835 \\
    \hline

    GunPointOldVersusYoung &  -     & 0.993651 & \textbf{1} & \textbf{1} & \textbf{1} & 0.990476 \\
    \hline

    HouseTwenty &  -     & 0.983193 & 0.983193 & 0.907563 & 0.983193 & \textbf{0.991597} \\
    \hline

    InsectEPGRegularTrain &  -     & 0.995984 & \textbf{1} & \textbf{1} & \textbf{1} & \textbf{1} \\
    \hline

    InsectEPGSmallTrain &  -     & 0.935743 & 0.955823 & 0.927711 & \textbf{0.971888} & 0.477912 \\
    \hline
 
    MelbournePedestrian &  -     & \textbf{0.913061} & 0.912245 & 0.911837 & 0.904898 & 0.901633 \\
    \hline

    MixedShapesRegularTrain &  -     & 0.973608 & 0.97567 & 0.969897 & 0.97567 & \textbf{0.980206} \\
    \hline

    MixedShapesSmallTrain &  -     & 0.936082 & 0.910103 & 0.918763 & 0.92866 & \textbf{0.940619} \\
    \hline

    PickupGestureWiimoteZ &  -     & 0.74  & \textbf{0.8} & 0.78  & 0.78  & 0.78 \\
    \hline
 
    PigAirwayPressure &  -     & \textbf{0.336538} & \textbf{0.336538} & 0.096154 & 0.173077 & 0.153846 \\
    \hline

    PigArtPressure & -      & \textbf{1} & \textbf{1} & 0.168269 & 0.043269 & 0.533654 \\
    \hline

    PigCVP &  -     & 0.875 & \textbf{0.908654} & 0.081731 & 0.211538 & 0.019231 \\
    \hline

    PLAID & -      & 0.901304 & 0.944134 & 0.921788 & 0.147114 & \textbf{0.945996} \\
    \hline

    PowerCons & -      & \textbf{0.994444} & 0.933333 & 0.944444 & 0.927778 & 0.927778 \\
    \hline

    Rock  &  -     & \textbf{0.92} & 0.78  & \textbf{0.92} & 0.82  & 0.76 \\
    \hline

    SemgHandGenderCh2 &   -    & 0.91  & 0.866667 & \textbf{0.916667} & 0.848333 & 0.651667 \\
    \hline

    SemgHandMovementCh2 &  -     & \textbf{0.56} & 0.513333 & 0.504444 & 0.391111 & 0.468889 \\
    \hline

    SemgHandSubjectCh2 & -      & \textbf{0.873333} & 0.746667 & 0.74  & 0.666667 & 0.788889 \\
    \hline

    ShakeGestureWiimoteZ &  -     & 0.88  & \textbf{0.94} & \textbf{0.94} & \textbf{0.94} & \textbf{0.94} \\
    \hline
    
    SmoothSubspace & -      & 0.98  & \textbf{1} & \textbf{1} & 0.993333 & \textbf{1} \\
    \hline
    
    UMD   & - & 0.986111 & \textbf{1} & \textbf{1} & \textbf{1} & \textbf{1} \\
    \hline

    \end{supertabular}}
 
\end{center}

\begin{table}[htbp]
  \centering
 \caption{Wilcoxon signed rank test comparison of each model}
  \resizebox{\textwidth}{25mm}{
  
        \begin{tabular}{|l|p{1.5cm}<{\centering}|c|c|c|c|p{1.5cm}<{\centering}|p{1.5cm}<{\centering}|c|p{1.5cm}<{\centering}|c|c|p{1.5cm}<{\centering}|p{1.5cm}<{\centering}|c|c|}
        \hline
          & resnet-\newline{}transformer & lstm-fcn & tfcn  & Resnet & fcn   & 1-NN DTW CV & 1-NN DTW & BOSS  & Learning \newline{} Shapelet (LS) & TSBF  & ST    & EE (PROP) & COTE (ensemble) & MLP   & CNN \\
          \hline
    lstm-fcn & 0.096581 & 0     &       &       &       &       &       &       &       &       &       &       &       &       &  \\
     \hline
    tfcn  & 3.76E-06 & 5.30E-06 & 0     &       &       &       &       &       &       &       &       &       &       &       &  \\
    \hline
    Resnet & 1.03E-08 & 1.82E-06 & 0.005163 & 0     &       &       &       &       &       &       &       &       &       &       &  \\
    \hline
    fcn   & 2.92E-11 & 1.45E-08 & 0.001401 & 0.958291 & 0     &       &       &       &       &       &       &       &       &       &  \\
    \hline
    1-NN DTW CV & 1.52E-13 & 3.48E-13 & 5.74E-12 & 3.69E-10 & 1.53E-07 & 0     &       &       &       &       &       &       &       &       &  \\
    \hline
    1-NN DTW & 2.81E-14 & 4.45E-14 & 5.64E-13 & 8.44E-13 & 2.98E-10 & 0.000381 & 0     &       &       &       &       &       &       &       &  \\
    \hline
    BOSS  & 1.79E-12 & 1.43E-10 & 2.52E-06 & 0.002485 & 0.012174 & 2.67E-07 & 1.11E-10 & 0     &       &       &       &       &       &       &  \\
    \hline
    Learning Shapelet (LS) & 7.05E-12 & 1.65E-11 & 8.39E-09 & 9.77E-06 & 0.000346 & 0.000416 & 1.04E-06 & 0.032497 & 0     &       &       &       &       &       &  \\
    \hline
    TSBF  & 1.49E-12 & 1.10E-12 & 1.65E-09 & 1.70E-08 & 1.59E-06 & 0.014259 & 5.39E-05 & 0.001402 & 0.423474 & 0     &       &       &       &       &  \\
    \hline
    ST    & 5.02E-11 & 8.09E-11 & 5.57E-06 & 0.006204 & 0.004078 & 1.61E-07 & 5.75E-10 & 0.15641 & 0.002506 & 0.00031 & 0     &       &       &       &  \\
    \hline
    EE (PROP) & 1.23E-11 & 7.22E-11 & 5.22E-08 & 1.66E-06 & 0.000117 & 8.67E-10 & 1.71E-11 & 0.022618 & 0.935015 & 0.518861 & 0.001094 & 0     &       &       &  \\
    \hline
    COTE (ensemble) & 2.38E-09 & 2.25E-07 & 0.002011 & 0.438745 & 0.144622 & 1.82E-13 & 5.82E-14 & 0.000584 & 2.20E-09 & 8.15E-09 & 0.001998 & 1.55E-11 & 0     &       &  \\
    \hline
    MLP   & 3.63E-13 & 4.91E-13 & 1.56E-12 & 5.94E-08 & 9.91E-07 & 0.924497 & 0.168452 & 0.000426 & 0.017383 & 0.046248 & 2.19E-05 & 0.016323 & 8.66E-08 & 0     &  \\
    \hline
    CNN   & 8.40E-09 & 4.20E-06 & 0.168655 & 0.222513 & 0.316416 & 9.73E-10 & 3.14E-12 & 0.001762 & 1.18E-06 & 4.24E-08 & 0.00088 & 8.24E-07 & 0.114445 & 2.78E-10 & 0 \\
    \hline
    
    \end{tabular}}
 
\end{table}%

\section{Conclusion}
We present RAN as a potential new architecture for time-series or sequence-driven modeling. Out of the standard 85 UCR data sets, we have achieved 35 state-of-the-art results with 10 results matching current state-of-the-art results without further model fine-tuning. The results indicate that such architecture is promising in complex, long-sequence modeling and may have vast, cross-domain applications.

\newpage 
\bibliographystyle{plain}
\bibliography{GraphStarNet}
\end{document}